\documentclass[10pt,twocolumn,letterpaper]{article}

\usepackage{cvpr}
\usepackage{times}
\usepackage{epsfig}
\usepackage{graphicx}
\usepackage{amsmath}
\usepackage{amssymb}
\usepackage{multirow}

\usepackage{bm}
\usepackage{booktabs} 
\usepackage{epstopdf}
\usepackage{url}
\usepackage{ulem}
\usepackage{color}
\usepackage{cite}

\usepackage{algorithm,algcompatible,amssymb,amsmath}

\algnewcommand\algorithmicto{\textbf{to}}
\algnewcommand\RETURN{\State \textbf{return} }

\algnewcommand\algorithmicinput{\textbf{Input:}}
\algnewcommand\INPUT{\item[\algorithmicinput]}

\algnewcommand\algorithmicoutput{\textbf{Output:}}
\algnewcommand\OUTPUT{\item[\algorithmicoutput]}

\algnewcommand\algorithmicinitialize{\textbf{Initialize:}}
\algnewcommand\INITIALIZE{\item[\algorithmicinitialize]}

\usepackage[breaklinks=true,bookmarks=false]{hyperref}

\cvprfinalcopy 

\def\cvprPaperID{****} 

\ifcvprfinal\fi

\usepackage{array}
\newcolumntype{L}[1]{>{\raggedright\let\newline\\\arraybackslash\hspace{0pt}}m{#1}}
\newcolumntype{C}[1]{>{\centering\let\newline\\\arraybackslash\hspace{0pt}}m{#1}}
\newcolumntype{R}[1]{>{\raggedleft\let\newline\\\arraybackslash\hspace{0pt}}m{#1}}
\begin{document}

\title{Object Region Mining with Adversarial Erasing: A Simple Classification to Semantic Segmentation Approach}


\author{\normalsize{Yunchao~Wei$^{1}$ \quad Jiashi~Feng$^{1}$ \quad Xiaodan Liang$^{2}$ \quad Ming-Ming~Cheng$^{3}$ \quad Yao~Zhao $^{4}$ \quad Shuicheng~Yan$^{1,5}$}\\
	\small{$^{1}$ National University of Singapore \quad $^{2}$ CMU \quad $^{3}$ Nankai University \quad $^{4}$ Beijing Jiaotong University \quad $^{5}$ 360 AI Institute} \\
	{\small  \{eleweiyv, elefjia\}@nus.edu.sg} \quad \small xiaodan1@cs.cmu.edu \quad  \small cmm@nankai.edu.cn \quad \small  yzhao@bjtu.edu.cn \small  \quad yanshuicheng@360.cn
}


\maketitle
\thispagestyle{empty}

\begin{abstract}

	We investigate a principle way to progressively mine discriminative object regions using classification networks to address the weakly-supervised semantic segmentation problems. Classification networks are only responsive to small and sparse discriminative regions from the object of interest, which deviates from the requirement of the segmentation task that needs to localize dense, interior and integral regions for pixel-wise inference. To mitigate this gap, we propose a new adversarial erasing approach for localizing and expanding object regions progressively. Starting with a single small object region, our proposed approach drives the classification network to sequentially discover new and complement object regions by erasing the current mined regions in an adversarial manner. These localized regions eventually constitute a dense and complete object region for learning semantic segmentation. To further enhance the quality of the discovered  regions by adversarial erasing, an online prohibitive segmentation learning approach is developed to collaborate with adversarial erasing by providing auxiliary segmentation supervision modulated by the more reliable classification scores. Despite its apparent simplicity, the proposed approach achieves 55.0\% and 55.7\%  mean Intersection-over-Union (mIoU) scores on PASCAL VOC 2012 val and test sets, which are the new state-of-the-arts.
	
\end{abstract}

\vspace{-3mm}
\section{Introduction}
\vspace{-2mm}
Deep neural networks (DNNs) have achieved remarkable success on semantic segmentation tasks~\cite{2015-long,chen2014semantic,zheng2015conditional,liu2015matching}, arguably benefiting from available resources of pixel-level annotated masks. However, collecting a large amount of accurate pixel-level annotation for training semantic segmentation networks on new image sets is labor intensive and inevitably requires substantial financial investments. To relieve the demand for the expensive pixel-level image annotations, \emph{weakly-supervised} approaches~\cite{liu2012weakly,pathak2014fully,2015-papandreou-weakly,pinheiro2015weakly,pathak2015constrained,kolesnikov2016seed,saleh2016built,qi2016augmented,shimoda2016distinct,russakovsky2015s,lin2016scribblesup,wei2015stc,wei2016learning} provide some promising solutions.
\begin{figure}[t]
	\centering
	\includegraphics[scale=0.46]{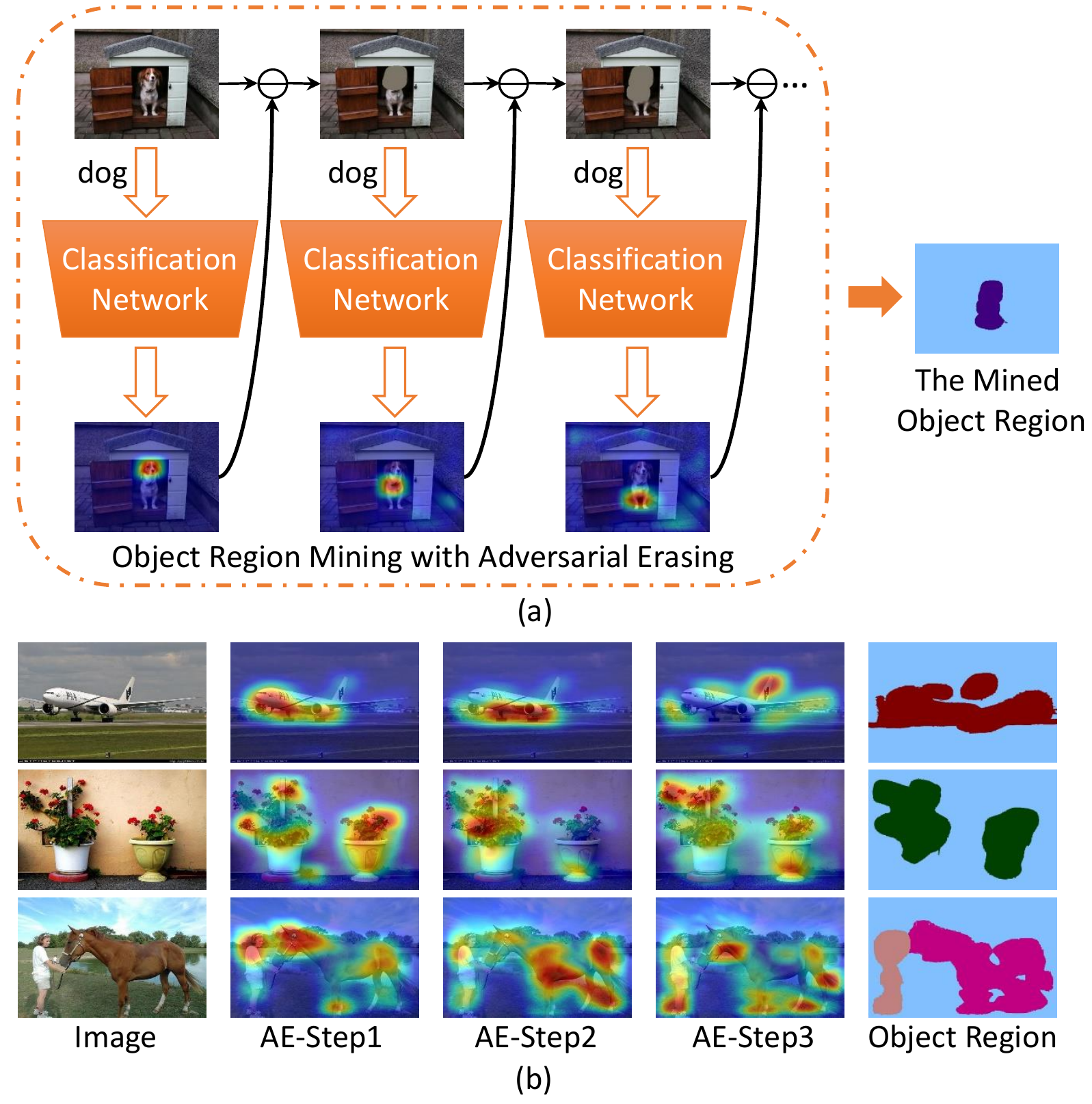}
	\caption{(a) Illustration of the proposed AE approach. With AE, a classification network first mines the most discriminative region for image category label ``dog". Then,  AE erases the mined region (\emph{head}) from the image and the classification network is re-trained to discover a new object region (\emph{body}) for performing classification without performance drop. We repeat such adversarial erasing process for multiple times and merge the erased regions into an integral foreground segmentation mask. (b) Examples of the discriminative object regions mined by AE at different steps and the obtained foreground segmentation masks in the end.}
	\label{fig:illu}
	\vspace{-3mm}
\end{figure}

Among various levels of weak supervision information, the simplest and most efficient one that can be collected for training semantic segmentation models is the image-level annotation~\cite{wei2015hcp,zhang2016online}. However, to train a well-performing semantic segmentation model given only such image-level annotation is rather challenging -- one obstacle is how to accurately assign image-level labels to corresponding pixels of training images such that DNN-based approaches can learn to segment images end-to-end. To establish the desired label-pixel correspondence, some approaches are developed that can be categorized as proposal-based and classification-based. The proposal-based methods~\cite{wei2016learning,qi2016augmented} often exhaustedly examine each proposal to generate pixel-wise masks, which are quite time-consuming. In contrast, the classification-based methods~\cite{kolesnikov2016seed,shimoda2016distinct,pathak2014fully,pinheiro2015weakly,pathak2015constrained,2015-papandreou-weakly} provide much more efficient alternatives. Those methods employ a classification model to select the regions that are most discriminative for the classification target and employ the regions as pixel-level supervision for semantic segmentation learning. However, object classification models usually identify and rely on a small and sparse discriminative region (as highlighted in the heatmaps produced by the classification network shown in Figure~\ref{fig:illu} (a)) from the object of interest. It deviates from requirement of the segmentation task that needs to localize dense, interior and integral regions for pixel-wise inference. Such deviation makes the main obstacle to adapting classification models for solving segmentation problems and harms the segmentation results. To address this issue, we propose a novel \emph{adversarial erasing} (AE) approach that is able to drive a classification network to learn integral object regions progressively. The AE approach can be viewed as establishing a line of competitors, trying to challenge the classification networks to discover some evidence of a specific category until no supportable evidence is left.


Concretely, we first train an image classification network using the image-level weak supervision information, \ie the object category annotation. The classification network is applied to localize the most discriminative region within an image for inferring the object category. We then erase the discovered region from the image to breakdown the performance of the classification network. To remedy the performance drop, the classification network needs to localize another discriminative region for classifying the image correctly. With such repetitive adversarial erasing operation, the classification network is able to mine other discriminative regions belonging to the object of interest. The process is illustrated by an example in Figure~\ref{fig:illu} (a), in which \emph{head} is the most discriminative part for classifying the ``dog" image. After erasing \emph{head} and re-training the classification network, another discriminative part \emph{body} would pop out. Repeating such adversarial erasing can localize increasingly discriminative regions diagnostic for image category until no more informative region left. Finally, the erased regions are merged to form a pixel-level semantic segmentation mask that can be used for training a segmentation model. More visualization examples are shown in Figure~\ref{fig:illu} (b).

However, the AE approach may miss some object-related regions and introduce some noise due to less attention on boundaries. To exploit those ignored object-related regions as well as alleviate noise, we further propose a complementary online \emph{prohibitive segmentation learning} (PSL) approach to work with AE together to discover more complete object regions and learn better semantic segmentation models. In particular, PSL uses the predicted image-level classification confidences to modulate the corresponding category-specific response maps and form them into an auxiliary segmentation mask, which can be updated in an online manner. Those category-specific segmentation maps with low classification confidences are prohibited for contributing to the formed supervision mask, thus noise can be reduced effectively.


To sum up, our main contributions are three-fold:
\vspace{-2mm}
\begin{itemize}
	\item We propose a new AE approach to effectively adapt an image classification network to continuously mining and expanding target object regions, and it eventually produces contiguous object segmentation masks that are usable for training segmentation models.
	\vspace{-2mm}
	\item We propose an online PSL method to utilize image-level classification confidences to reduce noise within the supervision mask and achieve better training of the segmentation network, collaborating with AE.
	\vspace{-2mm}
	\item Our work achieves the mIoU 55.0\%  and 55.7\% on \emph{val} and \emph{test} of the PASCAL VOC segmentation benchmark respectively, which are the new state-of-the-arts. 
	\vspace{-5mm}
\end{itemize}

\begin{figure*}[t]
	\centering
	\includegraphics[scale=0.500]{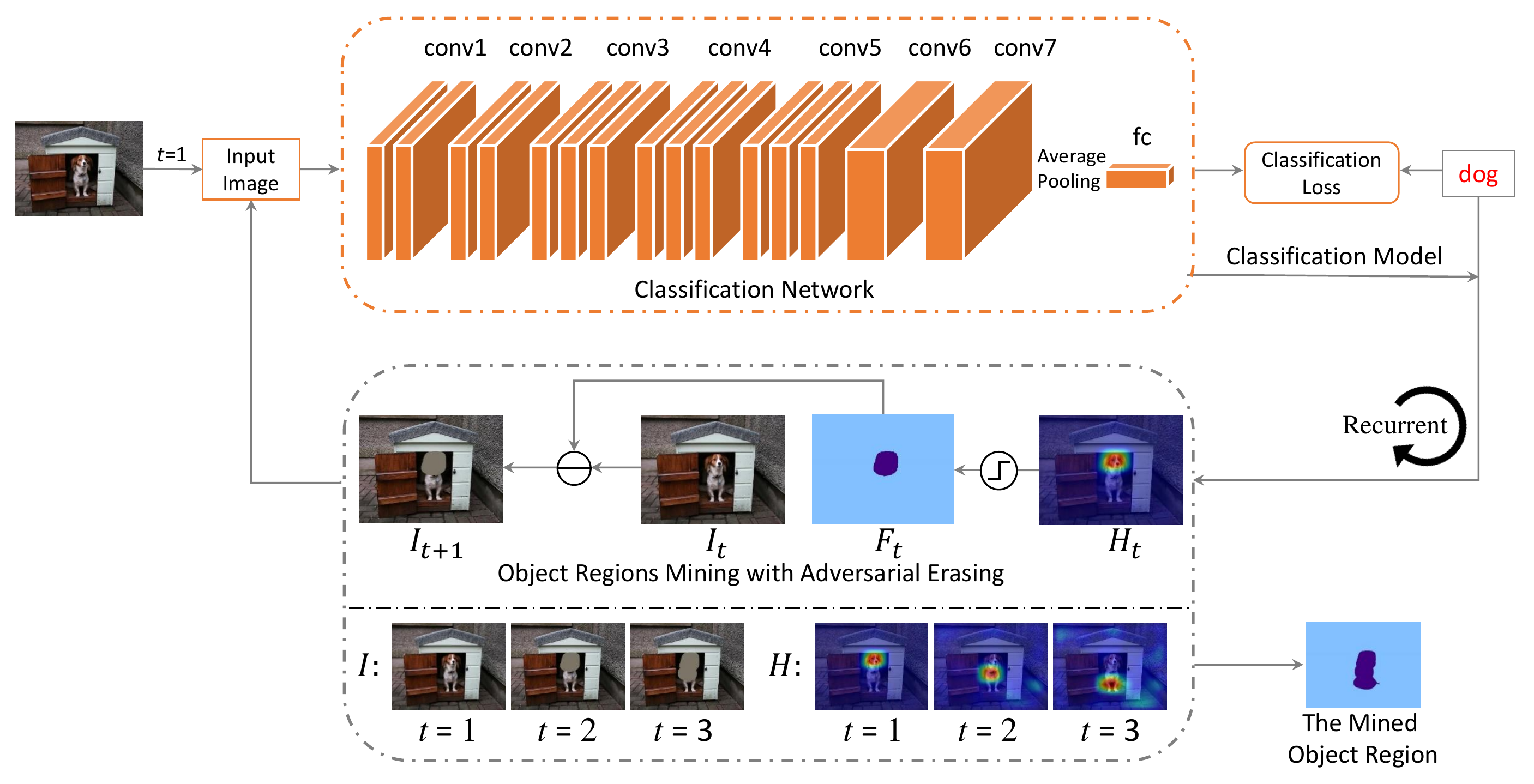}
	\caption{Overview of the proposed adversarial erasing approach. At the step $t$, we first train the classification network with the current processed image $I_t$; then a classification activation method (\eg CAM~\cite{zhou2015cnnlocalization}) is employed to produce the class-specific response heatmap ($H_t$). Applying hard thresholding on the heatmap $H_t$ reveals the discriminative region $F_t$. The proposed approach then erases $F_t$ from $I_t$ and produces $I_{t+1}$. This image is then fed into the classification network for learning to localize a new discriminative region. The learned heatmaps and corresponding proceeded training images with erasing are shown in the bottom. The mined regions from multiple steps together constitute the predicted object regions as output, which is used for training the segmentation network later.}
	\label{fig:mining}
	\vspace{-1.5em}
\end{figure*}

\section{Related Work}

To reduce the burden of pixel-level annotation, various weakly-supervised methods have been proposed for learning to perform semantic segmentation with coarser annotations. For example, Papandreou \etal~\cite{2015-papandreou-weakly} and Dai \etal~\cite{2015-dai} proposed to estimate segmentation using annotated bounding boxes. More recently, Lin \etal~\cite{lin2016scribblesup} employed scribbles as supervision for semantic segmentation. In~\cite{russakovsky2015s}, the required supervised information is further relaxed to instance points. All these annotations can be considered much simpler than pixel-level annotation. 

Some works~\cite{pathak2014fully,2015-papandreou-weakly,pinheiro2015weakly,pathak2015constrained,vezhnevets2011weakly,xu2015learning} propose to train the segmentation models by only using image-level labels, which is the simplest supervision for training semantic segmentation models. Among those works, Pinheiro \etal~\cite{pinheiro2015weakly} and Pathak \etal~\cite{pathak2014fully} proposed to utilize multiple instance learning (MIL) to train the models for segmentation. Pathak \etal~\cite{pathak2015constrained} introduced a constrained CNN model to address this problem. Papandreou \etal~\cite{2015-papandreou-weakly} adopted an alternative training procedure based on the Expectation-Maximization algorithm to dynamically predict semantic foreground and background pixels. However, the performance of those methods is not satisfactory. Recently, some new approaches~\cite{kolesnikov2016seed,saleh2016built,qi2016augmented,shimoda2016distinct,wei2015stc,wei2016learning} are proposed to further improve the performance of this challenging task. In particular, Wei \etal~\cite{wei2015stc} presented a simple to complex learning method, in which an initial segmentation model is trained with simple images using saliency maps for supervision. Then, samples of increasing complexity are progressively included to further enhance the ability of the segmentation model. In~\cite{kolesnikov2016seed}, three kinds of loss functions, \ie seeding, expansion and constrain-to-boundary, are proposed and integrated into a unified framework to train the segmentation network. Both~\cite{kolesnikov2016seed} and our work propose to localize object cues according to classification networks. However, Kolesnikov \etal~\cite{kolesnikov2016seed} can only obtain small and sparse object-related seeds for supervision. In contrast, the proposed AE approach is able to mine dense object-related regions, which can provide richer supervised information for learning to perform semantic segmentation. In addition, Qi \etal~\cite{qi2016augmented} proposed an augmented feedback method, in which GrabCut~\cite{rother2004grabcut} and object proposals are employed to generate pixel-level annotations for supervision. To the best of our knowledge, Qi \etal~\cite{qi2016augmented} achieved the state-of-the-art mIoU scores using Selective Search~\cite{uijlings2013selective} (52.7\%) and MCG~\cite{arbelaez2014multiscale} (55.5\%) segmentation proposals on the PASCAL VOC benchmark. However, note that MCG has been trained from PASCAL \emph{train} images with pixel-level annotations, and thus the corresponding results of \cite{qi2016augmented} are obtained by using stronger supervision inherently.

\vspace{-0.5em}
\section {Classification to Semantic Segmentation}
The proposed classification to semantic segmentation approach includes two novel components, \ie object region mining with AE and online PSL for semantic segmentation.


\subsection{Object Region Mining with AE}
To address the problem that classification networks are only responsive to small and sparse discriminative regions, we propose the AE approach for localizing and expanding object regions progressively.
As shown in Figure~\ref{fig:mining}, the AE iteratively performs two operations: learning a classification network for localizing the object discriminative regions and adversarially erasing the discovered regions. In particular, the classification network is initialized based on the DeepLab-CRF-LargeFOV~\cite{chen2014semantic} model. Global average pooling is applied on \emph{conv7} and the generated representations pass through a fully-connected layer for predicting classification. In the first operation, we train the classification network by minimizing squared label prediction loss as suggested by~\cite{wei2015hcp}. In the second operation of performing erasing, we first produce the heatmap for each image-level label using the classification activation maps (CAM) method~\cite{zhou2015cnnlocalization}. Then, the discriminative object regions are obtained by applying a hard threshold to the heatmap. We erase the mined region from training images by replacing its internal pixels by the mean pixel values of all the training images. The processed image with erased regions is then fed into the next classification learning iteration. As the discriminative regions have been removed and no longer contribute to the classification prediction, the classification network is naturally driven to discover new object discriminative regions for maintaining its classification accuracy level. We repeat the classification learning and the AE process for several times until the network cannot well converge on the produced training images, \ie no more discriminative regions left for performing reasonably good classification.

We now explain the AE process more formally. Suppose the training set $\mathcal{I}=\{(I_i, \mathcal{O}_i)\}_{i=1}^N$ includes $N$ images and $\mathcal{F}=\{F_i\}_{i=1}^N$ represents the mined object regions by AE. We iteratively produce the object regions $F_{i,t}$ for each training image $I_{i,t}$ with the classification model $M_t$ at the $t^{th}$ learning step. Denote $\mathcal{C}$ as the set of object categories and CAM$(\cdot)$ as the operation of heatmap generation. Thus, the $c^{th}$ heatmap $H_{i,t}^c$ of $I_{i,t}$, in which $c \in \mathcal{O}_i$ and $\mathcal{O}_i \subseteq \mathcal{C}$ is the image-level label set of $I_{i,t}$, can be obtained according to CAM$(I_{i,t}, M_t, c)$. To enforce the classification network to expand object regions from $I_{i,t}$, we erase the pixels whose values on $H_{i,t}^c$ are larger than $\delta$. Then, $\mathcal{F}$ is obtained through the procedure summarized in Algorithm 1. 

\begin{algorithm}[t]
	\caption{Object Regions Mining with AE}
	\label{algo}
	\begin{algorithmic}[1]
		\INPUT \text{Training data} $\mathcal{I}=\{(I_i, \mathcal{O}_i)\}_{i=1}^N$, \text{threshold} $\delta$.
		\INITIALIZE $F_i=\varnothing (i=1,\cdots,N)$, $t=1$.
		\WHILE {(training of classification is success)}
		\STATE Train the classification network $M_t$ with $\mathcal{I}$.
		\FOR{$I_i$ in $\mathcal{I}$}
		\STATE {{\bfseries Set} $F_{i,t}=\varnothing$.}
		\FOR{$c$ in $\mathcal{O}_i$}
		\STATE{Calculate $H_{i,t}^c$ by CAM$(I_{i,t}, M_t, c)$~\cite{zhou2015cnnlocalization}.}
		\STATE{Extract regions $R$ whose corresponding pixel  \phantom .\phantom . \phantom .\phantom . \phantom . \phantom . \phantom .\phantom . \phantom . values in $H_{i,t}^c$ are larger than $\delta$.}
		\STATE{Update the mined regions $F_{i,t}^c= F_{i,t}^c \cup R$.}
		\ENDFOR
		\STATE{Update the mined regions $F_i= F_i \cup F_{i,t}$.}
		\STATE{Erase the mined regions from training image \phantom . \phantom .\phantom . \phantom .\phantom . \phantom . \phantom . \phantom .\phantom . \phantom .$I_{i,{t+1}}=I_{i,t} \backslash F_{i,t}$.}
		\ENDFOR
		\STATE{$t = t + 1 $.}	
		
		\ENDWHILE
		\OUTPUT $\mathcal{F}=\{F_i\}_{i=1}^N$
		
	\end{algorithmic}
	
\end{algorithm}


Beyond mining foreground object regions, finding background localization cues is also crucial for training the segmentation network. Motivated by~\cite{wei2015stc,kolesnikov2016seed}, we use the saliency detection technology~\cite{jiang2013salient} to produce the saliency maps of training images. Based on the generated saliency maps, the regions whose pixels are with low saliency values are selected as background. Suppose $B_i$ denotes the selected background regions of $I_i$. We can obtain the segmentation masks  $\mathcal{S}=\{S_i\}_{i=1}^N$, where $S_i = F_i \cup B_i$. We ignore three kinds of pixels for producing $\mathcal{S}$: 1) those erased foreground regions of different categories which are in conflict; 2) those
low-saliency pixels which lie within the object regions identified by AE; 3) those pixels that are not assigned semantic labels. One example of the segmentation mask generation process is demonstrated in Figure 3 (a). ``black" and ``purple" regions refer to the background and the object, respectively.
\begin{figure}[t]
	\centering
	\includegraphics[scale=0.5]{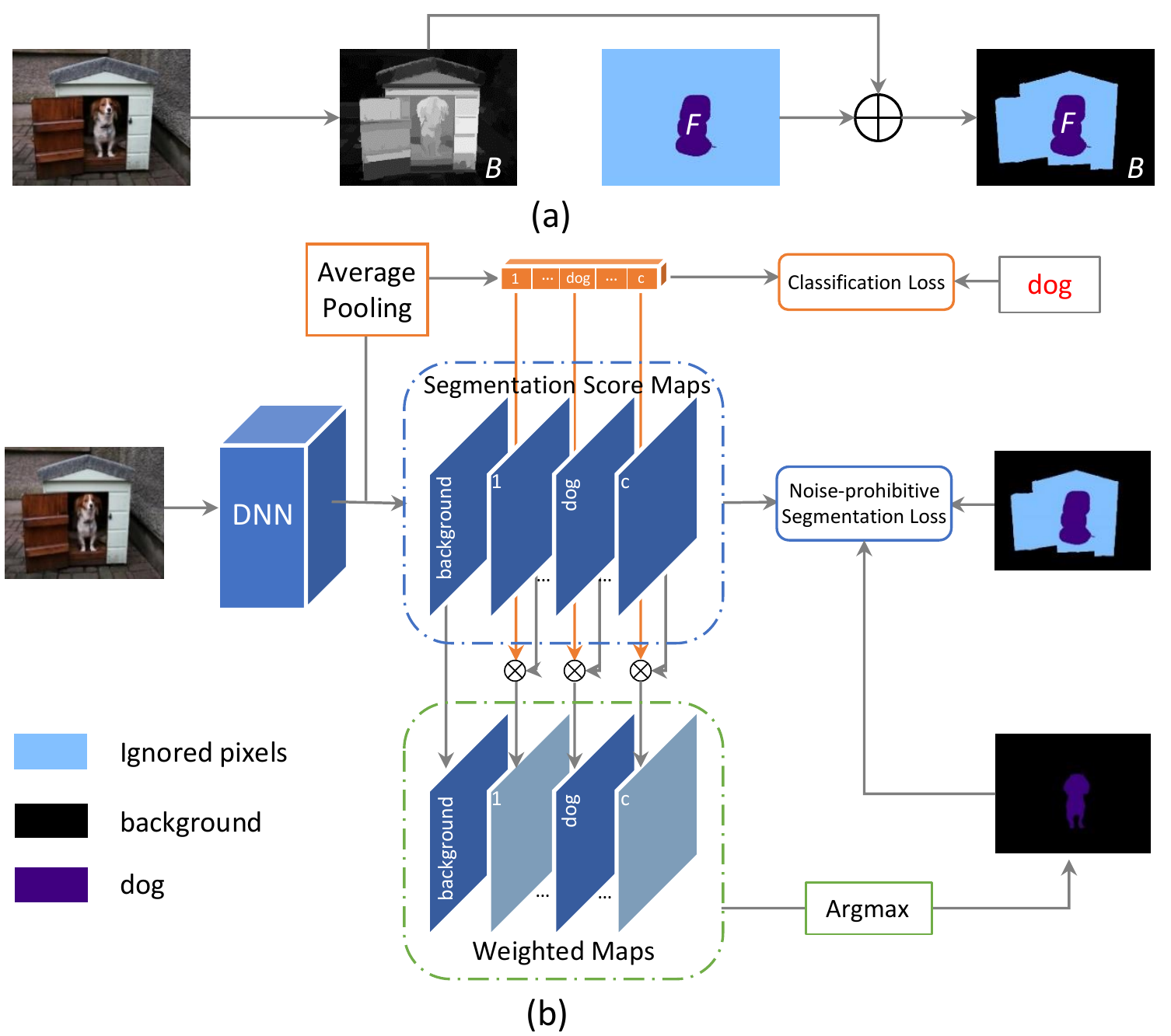}
	\caption{(a) The process of segmentation mask generation. (b) The proposed online PSL approach for semantic segmentation. The classification scores are used to weight ``Segmentation Score Maps" to produce ``Weighted Maps" in an online manner. Those classes with low classification confidences are prohibited for producing the segmentation mask. Then, both the mined mask and the online produced mask are used to optimize the network. }
	\label{fig:training}
	\vspace{-2em}
\end{figure}

\subsection{Online PSL for Semantic Segmentation}
The proposed AE approach provides the initial segmentation mask for each training image that can be used for training segmentation networks. However, some object-related or background-related pixels may be missed (as those ``blue" pixels on the AE outputs shown in Figure~\ref{fig:training} (a)). In addition, semantic labels of some labeled pixels may be noisy due to the limitation of AE on capturing boundary details. To exploit those pixels unlabeled by AE for training and gain robustness to falsely labeled pixels, we propose an online Prohibitive Segmentation Learning (PSL) approach to further learn to perform semantic segmentation upon the masks provided by AE. The online PSL exploits image classification results to identify reliable category-wise segmentation maps and form them into a less noisy auxiliary supervision map, offering auxiliary information to the AE output. PSL updates the produced auxiliary segmentation map along with training of the segmentation networks in an online manner and produces increasingly more reliable auxiliary supervision. As shown in Figure~\ref{fig:training} (b), the proposed PSL builds a framework that includes two branches, one for classification and the other for semantic segmentation. In particular, PSL uses the squared loss as the optimization objective for the classification branch, whose produced classification confidences are used by PSL to weight the corresponding category-specific segmentation score maps. With the help of classification results, the online PSL is able to integrate the multi-category segmentation maps into an auxiliary segmentation mask and provides supervision in addition to the AE output. With PSL, those segmentation maps corresponding to categories with low classification confidences are prohibited from contributing to the auxiliary segmentation map. Thus, noise from those irrelevant categories can be effectively alleviated.

Formally, denote the set of semantic labels for segmentation task as $\mathcal{C}^{seg}$ and the image-specific label set for a given image $I$ as $\mathcal{O}^{seg}$, in which background category is included. During each training epoch, we denote the image-level prediction from the classification branch as $\bm{v}$. Suppose $S$ is the segmentation mask produced by AE. The online PSL exploits the image prediction over $\mathcal{C}^{seg}$ to train a segmentation network $f(I;\theta)$ parameterized by $\theta$, which predicts the pixel-wise probability of each label $c \in \mathcal{C}^{seg}$ at every location $u$ of the image plane $f_{u,c}(I, \theta)$. To produce the additional segmentation mask $\hat{S}$ for training the segmentation network, PSL uses $\bm{v}$ to weight foreground category segmentation score maps as shown in Figure~\ref{fig:training} (b). With this prohibitive operation, large response values from negative score maps can be suppressed by multiplying a small classification category score. Meanwhile, the score maps of dominant categories (\ie the corresponding objects that occupy a large area of the image) can also be enhanced. Denote the weighting operator as $\otimes$, and $\hat{S}$ is then produced by
\vspace{-0.8em}
\[
\hat{S} = \mathop {\max } \{{[1, \bm{v}] \otimes f(I;\theta)}\}.\vspace{-0.5em}
\]
Here the appended element 1 is for weighting the background category. Suppose $S_c$ and $\hat{S}_c$ represent the pixels annotated with category $c$. The cross-entropy loss used for noise-prohibitive semantic segmentation is formulated as
\vspace{-0.5em}
\[
\min\limits_{\theta}\sum\limits_{I \in \mathcal{I}}{J(f(I;\theta), S) + J(f(I;\theta), \hat{S})} \vspace{-0.5em} \vspace{-0.5em}
\]
where 
\vspace{-0.8em}
\[
\label{eq:loss_seg}
J(f(I;\theta), S) = -{\frac{1}{\sum\limits_{c \in \mathcal{O}^{seg}}|S_c|} }\sum\limits_{c \in \mathcal{O}^{seg}}\sum\limits_{u \in S_{c}}\log f_{u,c}(I;\theta),\vspace{-0.5em}
\]
and
\vspace{-0.5em}
\[
\label{eq:loss_seg2}
J(f(I;\theta), \hat{S}) = -{\frac{1}{\sum\limits_{c \in \mathcal{O}^{seg}}|\hat{S}_c|} }\sum\limits_{c \in \mathcal{O}^{seg}}\sum\limits_{u \in \hat{S}_{c}}\log f_{u,c}(I;\theta).\vspace{-0.5em}
\]
With online training, the segmentation ablity of the network is progressively improved, which can produce increasingly more accurate $\hat{S}$ for supervising the later training process.

During the testing process, we take a more strict pohibitive policy for those categories with low classification confidences. In particular, we set those classification confidences that are smaller than $p$ to zero and keep others unchanged, and apply them to weight the predicted segmentation score maps and produce the final segmentation result.

\vspace{-0.5em}
\section{Experiments}
\label{sec:exp}
\subsection{Dataset and Experiment Settings}
\noindent \textbf{Dataset and Evaluation Metrics} We evaluate our proposed approach on the PASCAL VOC 2012 segmentation benchmark dataset~\cite{2010-pascal}, which has 20 object categories and one background category. This dataset is split into three subsets: training (\emph{train}, 1,464 images), validation (\emph{val}, 1,449 images) and testing (\emph{test}, 1,456 images). Following the common practice~\cite{pinheiro2015weakly,chen2014semantic,hariharan2011semantic}, we increase the number of training images to 10,582 by image augmentation. In our experiments, only image-level labels are utilized for training. The performance is evaluated in terms of pixel IoU averaged on 21 categories. Experimental analysis of the proposed approach is conducted on the \emph{val} set. We compare our method with other state-of-the-arts on both \emph{val} and \emph{test} sets. The result on the \emph{test} set is obtained by submitting the predicted results to the official PASCAL VOC evaluation server.

\noindent \textbf{Training/Testing Settings } We adopt DeepLab-CRF-LargeFOV from \cite{chen2014semantic} as the basic network for the classification network and segmentation network in AE and PSL, whose parameters are initialized by the VGG-16~\cite{simonyan2014very} pre-trained on ImageNet~\cite{2009-imagenet}. We use a mini-batch size of 30 images where patches of 321 $\times$ 321 pixels are randomly cropped from images for training the network. We follow the training procedure in~\cite{chen2014semantic} at this stage. The initial learning rate is 0.001 (0.01 for the last layer) and decreased by a factor of 10 after 6 epochs. Training terminates after 15 epochs. Both two networks are trained on NVIDIA GeForce TITAN X GPU with 12GB memory. We use DeepLab code~\cite{chen2014semantic} in our experiments, which is implemented based on the publicly available Caffe framework~\cite{jia2014caffe}. 

For each step of AE, those pixels belonging to top 20\% of the largest value (a fraction suggested by~\cite{kolesnikov2016seed,zhou2015cnnlocalization}) in the heatmap are erased, which are then considered as foreground object regions. We use saliency maps from \cite{jiang2013salient} to produce the background localization cues. For those images belonging to indoor scenes (\eg \emph{sofa} or \emph{table}), we adopt the normalized saliency value 0.06 as the threshold to obtain background localization cues (\ie pixels whose saliency values are smaller than 0.06 are considered as background) in case some objects were wrongly assigned to background. For the images from other categories, the threshold is set as 0.12. For the testing phase of semantic segmentation, the prohibited threshold $p$ is empirically set as 0.1 and CRF~\cite{koltun2011efficient} is utilized for post processing.

\begin{table}[]
	\centering
	\caption{Comparison of weakly-supervised semantic segmentation methods on VOC 2012 \emph{val} set.}
	\small
	\label{tab:val-comp}
	\begin{tabular}{lcc}
		\toprule
		Methods & Training Set & mIoU  \\
		\midrule
		\multicolumn{2}{l}{Supervision: Scribbles}  \\
		Scribblesup (CVPR 2016)~\cite{lin2016scribblesup} & 10K & 63.1 \\
		\midrule
		\multicolumn{2}{l}{Supervision: Box}  \\
		WSSL (ICCV 2015)~\cite{2015-papandreou-weakly} & 10K & 60.6 \\
		BoxSup (ICCV 2015) & 10K & 62.0 \\		
		\midrule
		\multicolumn{2}{l}{Supervision: Spot}  \\
		1 Point (ECCV 2016)~\cite{russakovsky2015s} & 10K & 46.1 \\
		Scribblesup (CVPR 2016)~\cite{lin2016scribblesup} & 10K & 51.6 \\		
		\midrule
		\multicolumn{2}{l}{Supervision: Image-level Labels}  \\	
		\multicolumn{3}{l}{(* indicates methods implicitly use pixel-level supervision)}\\
		SN\_B* (PR 2016)~\cite{wei2016learning} & 10K & 41.9\\
		MIL-seg* (CVPR 2015)~\cite{pinheiro2015weakly} & 700K & 42.0 \\
		TransferNet* (CVPR 2016)~\cite{hong2015learning} & 70K & 52.1 \\
		AF-MCG* (ECCV 2016)~\cite{qi2016augmented} & 10K &  54.3\\
		\midrule
		\multicolumn{2}{l}{Supervision: Image-level Labels}  \\		
		MIL-FCN (ICLR 2015)~\cite{pathak2014fully} & 10K & 25.7 \\
		CCNN (ICCV 2015)~\cite{pathak2015constrained} & 10K & 35.3\\
		MIL-sppxl (CVPR 2015)~\cite{pinheiro2015weakly}  & 700K & 36.6 \\
		MIL-bb (CVPR 2015)~\cite{pinheiro2015weakly} & 700K & 37.8 \\
		EM-Adapt (ICCV 2015)~\cite{2015-papandreou-weakly} & 10K & 38.2\\								
		DCSM (ECCV 2016)~\cite{shimoda2016distinct} & 10K & 44.1\\
		BFBP (ECCV 2016)~\cite{saleh2016built} & 10K & 46.6\\
		STC (PAMI 2016)~\cite{wei2015stc} & 50K & 49.8 \\
		SEC (ECCV 2016)~\cite{kolesnikov2016seed} & 10K & 50.7 \\				
		AF-SS (ECCV 2016)~\cite{qi2016augmented} & 10K &  52.6\\		
		\midrule
		\multicolumn{2}{l}{Supervision: Image-level Labels}  \\			
		AE-PSL (ours) & 10K & $\bm{55.0}$ \\
		\bottomrule
		
	\end{tabular}
	\vspace{-2em}
\end{table}

\begin{table}[t]
	\centering
	\caption{Comparison of weakly-supervised semantic segmentation methods on VOC 2012 \emph{test} set.}
	\small
	\label{tab:test-comp}
	\begin{tabular}{lcc}
		\toprule
		Methods & Training Set & mIoU  \\
		\midrule
		\multicolumn{2}{l}{Supervision: Box}  \\
		WSSL (ICCV 2015)~\cite{2015-papandreou-weakly} & 10K & 62.2 \\
		BoxSup (ICCV 2015)~\cite{2015-dai} & 10K & 64.2 \\		
		\midrule
		\multicolumn{2}{l}{Supervision: Image-level Labels}  \\	
		\multicolumn{3}{l}{(* indicates methods implicitly use pixel-level supervision)} \\
		MIL-seg* (CVPR 2015)~\cite{pinheiro2015weakly} & 700K & 40.6 \\
		SN\_B* (PR 2016)~\cite{wei2016learning} & 10K & 43.2\\	
		TransferNet* (CVPR 2016)~\cite{hong2015learning} & 70K & 51.2 \\
		AF-MCG* (ECCV 2016)~\cite{qi2016augmented} & 10K &  55.5\\
		\midrule
		\multicolumn{2}{l}{Supervision: Image-level Labels}  \\			
		MIL-FCN (ICLR 2015)~\cite{pathak2014fully} & 10K & 24.9 \\
		CCNN (ICCV 2015)~\cite{pathak2015constrained} & 10K & 35.6\\
		MIL-sppxl (CVPR 2015)~\cite{pinheiro2015weakly}  & 700K & 35.8 \\
		MIL-bb (CVPR 2015)~\cite{pinheiro2015weakly} & 700K & 37.0 \\
		EM-Adapt (ICCV 2015)~\cite{2015-papandreou-weakly} & 10K & 39.6\\											
		DCSM (ECCV 2016)~\cite{shimoda2016distinct} & 10K & 45.1\\
		BFBP (ECCV 2016)~\cite{saleh2016built} & 10K & 48.0\\
		STC (PAMI 2016)~\cite{wei2015stc} & 50K & 51.2 \\		
		SEC (ECCV 2016)~\cite{kolesnikov2016seed} & 10K & 51.7 \\				
		AF-SS (ECCV 2016)~\cite{qi2016augmented} & 10K &  52.7\\	
		\midrule
		\multicolumn{2}{l}{Supervision: Image-level Labels}  \\			
		AE-PSL (ours) & 10K & $\bm{55.7}$ \\
		
		\bottomrule
	\end{tabular}
	\vspace{-0.5em}
\end{table}
\subsection{Comparisons with State-of-the-arts}
We make extensive comparisons with state-of-the-art weakly-supervised semantic segmentation solutions with different levels of annotations, including scribbles, bounding boxes, spots and image-level labels. Results of those methods as well as ours on PASCAL VOC \emph{val} are summarized in Table~\ref{tab:val-comp}. 
Among the baselines, MIL-*~\cite{pinheiro2015weakly}, STC~\cite{wei2015stc} and TransferNet~\cite{hong2015learning} use more images (700K, 50K and 70K) for training. All the other methods are based on 10K training images and built on top of the VGG16~\cite{simonyan2014very} model. 

From the result, we can observe that our proposed approach outperforms all the other works using image-level labels and point annotation for weak supervision. In particular, AF-MCG~\cite{qi2016augmented} achieves the second best performance among the baselines only using image-level labels. However, the MCG generator is trained in a fully-supervised way on PASCAL VOC, thus the corresponding result, \ie AF-MCG~\cite{qi2016augmented}, implicitly makes use of stronger supervision. Thus, with the Selective Search segments, the performance of AF-SS~\cite{qi2016augmented} drops by 1.7\%. Furthermore, GrabCut~\cite{rother2004grabcut} is also employed by AF-*\cite{qi2016augmented} to refine the segmentation masks for supervision, which is usually time consuming for training. In contrast, the proposed AE approach is very simple and convenient to carry out for object region mining. In addition, the online PSL is also effective and efficient for training the semantic segmentation network. Compared with those methods using image-level labels for supervision, the proposed AE-PSL improves upon the best performance by over 2.4\%. Besides, our approach also outperforms those methods that implicitly use pixel-level supervision by over 0.7\%. Additional comparison among these approaches on PASCAL VOC \emph{test} is shown in Table~\ref{tab:test-comp}. It can be seen that our method achieves the new state-of-the-art for this challenging task on a competitive benchmark.

\begin{figure}[t]
	\centering
	\includegraphics[scale=0.60]{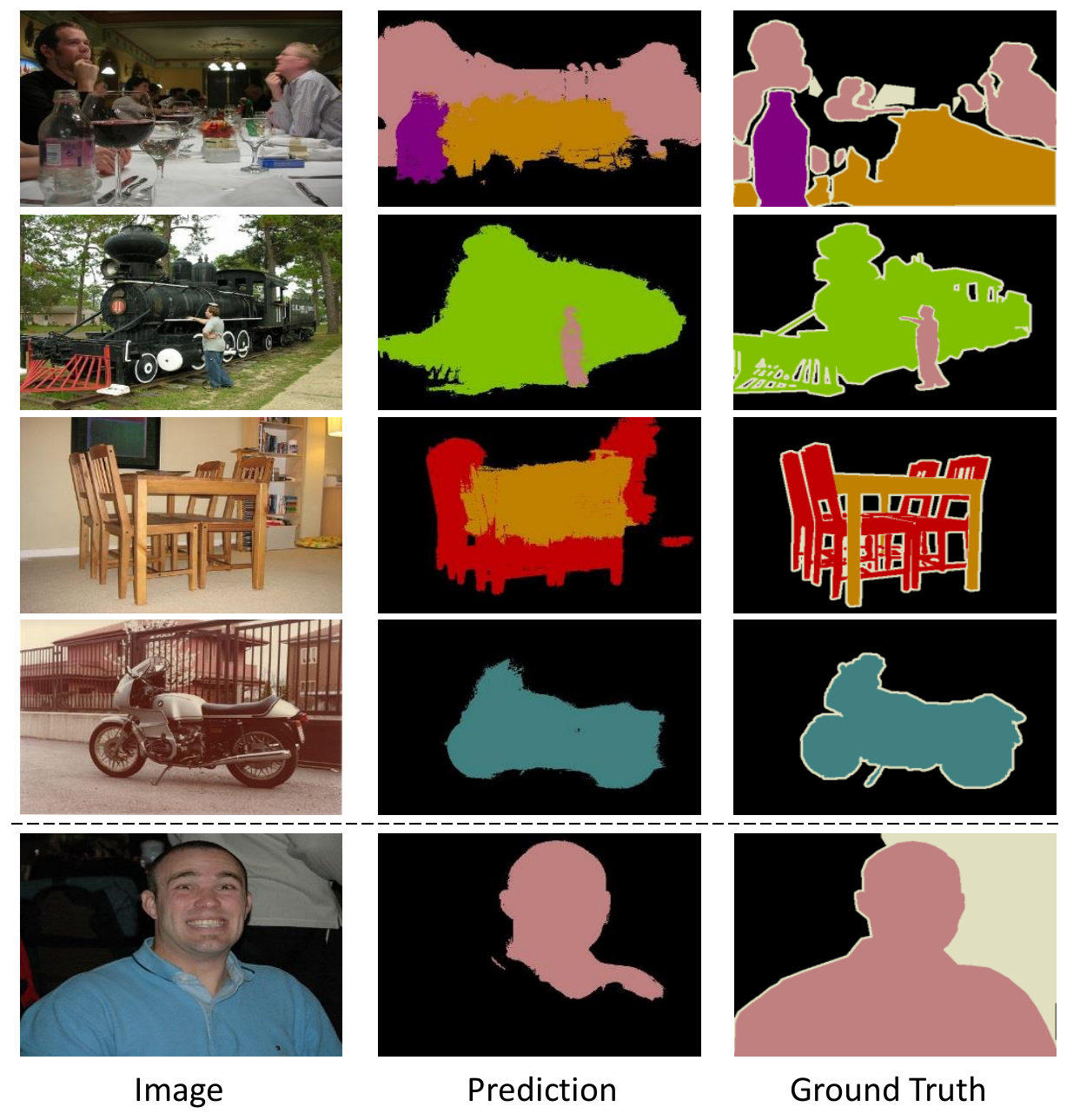}
	\caption{Qualitative segmentation results on the VOC 2012 \emph{val} set. One failure case is shown in the last row.}
	\label{fig:example}
	\vspace{-1em}
\end{figure}

Figure~\ref{fig:example} shows some successful segmentations, indicating that our method can produce accurate results even for some complex images. One typical failure case is given in the bottom row of Figure~\ref{fig:example}. This case may be well addressed with a better erasing strategy such as using low level visual features (\eg color and texture) to refine and extend erasing regions.

%
%

\begin{table*}\setlength{\tabcolsep}{2pt}
	\centering
	\caption{Comparison of segmentation mIoU scores using object regions from different AE steps on VOC 2012 \emph{val} set.}
	\label{tab:step}
	\footnotesize
	\begin{tabular}{C{1.5cm}|ccccccccccccccccccccc|c}
		\toprule
		AE Steps  & bkg & plane & bike  & bird  & boat  & bottle & bus   & car   & cat   & chair & cow   & table & dog   & horse & motor & person & plant & sheep & sofa  & train & tv &  mIoU  \\
		\midrule
		AE-step1 & 82.6 & 63.0 & 27.5 & 45.9 & 38.3 & 43.6 & 61.3 & 29.2 & 60.0 & 13.6 & 52.0 & 32.6 & 52.4 & 49.8 & 47.9 & 43.7 & 32.6 & 61.4 & 29.4 & 35.1 & 41.9 & 44.9\\
		AE-step2 & 82.2 & 69.3 & 29.7 & 60.9 & 40.8 & 52.4 & 59.3 & 44.2 & 65.3 & 13.0 & 58.9 & 32.2 & 60.0 & 56.6 & 49.1 & 43.0 & 34.2 & 69.7 & 32.1 & 42.8 & 43.2 & 49.5\\
		AE-step3 & 78.5 & 71.8 & 29.2 & 64.1 & 39.9 & 57.8 & 58.5 & 54.5 & 63.0 & 10.3 & 60.5 & 36.0 & 61.6 & 56.1 & 62.6 & 42.9 & 36.5 & 64.5 & 31.5 & 49.5 & 38.7 & 50.9\\
		\midrule
		AE-step4 & 74.4 & 65.5 & 28.2 & 59.7 & 38.5 & 57.8 & 57.5 & 59.0 & 57.2 & 9.6 & 54.9 & 39.2 & 56.5 & 52.6 & 65.0 & 43.2 & 34.9 & 55.9 & 30.4 & 47.9 & 36.8 & 48.8\\
		\bottomrule
	\end{tabular}
	\vspace{-2em}
\end{table*}

\begin{figure}[t]
	\centering
	\includegraphics[scale=0.65]{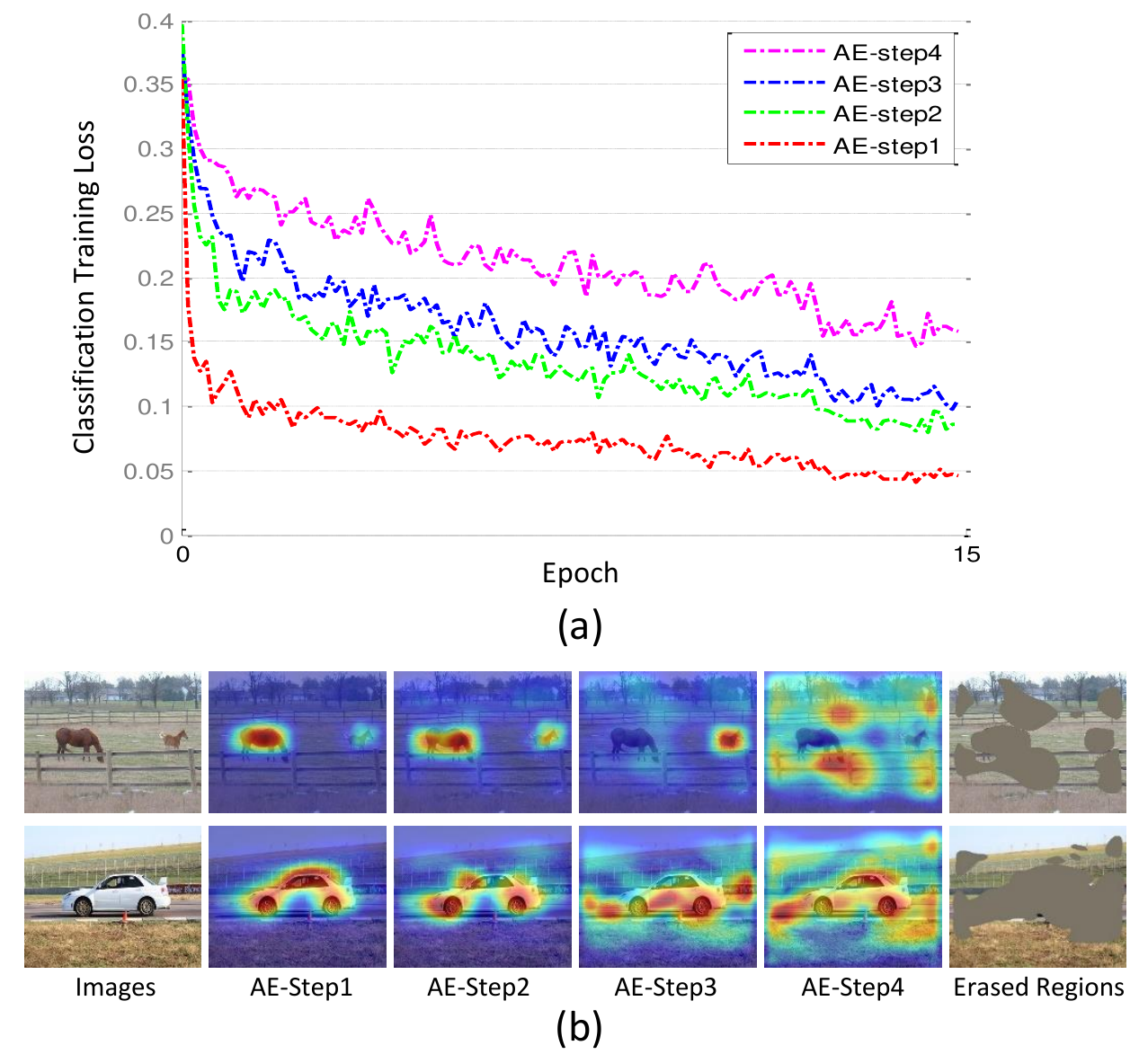}
	\caption{(a) Loss curves of classification network against varying numbers of training epochs, for different AE steps. (b) Failure cases of over erasing samples with four AE steps.}
	\label{fig:loss-failure}
	\vspace{-1.5em}
\end{figure}

\subsection{Ablation Analysis}
\vspace{-0.5em}
\subsubsection{Object Region Mining with AE}
\vspace{-0.5em}
\label{subsubsec:mdor}

With the AE approach, discriminative object regions are adversarially erased step by step. Therefore, it is expected that the loss values of the classification networks at the convergence of training across different AE steps would progressively increase as more discriminative regions are absent for training the classification networks. Figure~\ref{fig:loss-failure} (a) shows the comparison of the classification training loss curves for different AE steps. It can be observed that the loss value at convergence of training with original images is around 0.05. By performing the AE for multiple steps, the converged loss value slightly increases (AE-step2: $\sim$0.08, AE-step3: $\sim$0.1) compared with that of the AE-step1. This demonstrates that AE removes regions with a descending discriminative ability. By continuing to perform the AE for more steps to remove more regions, the classification network only converges to one that provides a training loss as large as $\sim$0.15. This demonstrates no more useful regions are left for obtaining a good classification network, due to \emph{over} \emph{erasing}. \emph{over} \emph{erasing} may introduce many true negative regions into the mined foreground object regions and hampers learning segmentation. Some failure cases caused by \emph{over} \emph{erasing} are shown in Figure~\ref{fig:loss-failure} (b). In the case where most object regions are removed from the training images, the classification network has to rely on some contextual regions to recognize the categories. These regions are true negative ones and detrimental for the segmentation network training. To prevent contamination from negative regions, we only integrate those discriminative regions mined from the first three steps into the final segmentation masks.

\begin{figure*}[t]
	\centering
	\includegraphics[scale=0.50]{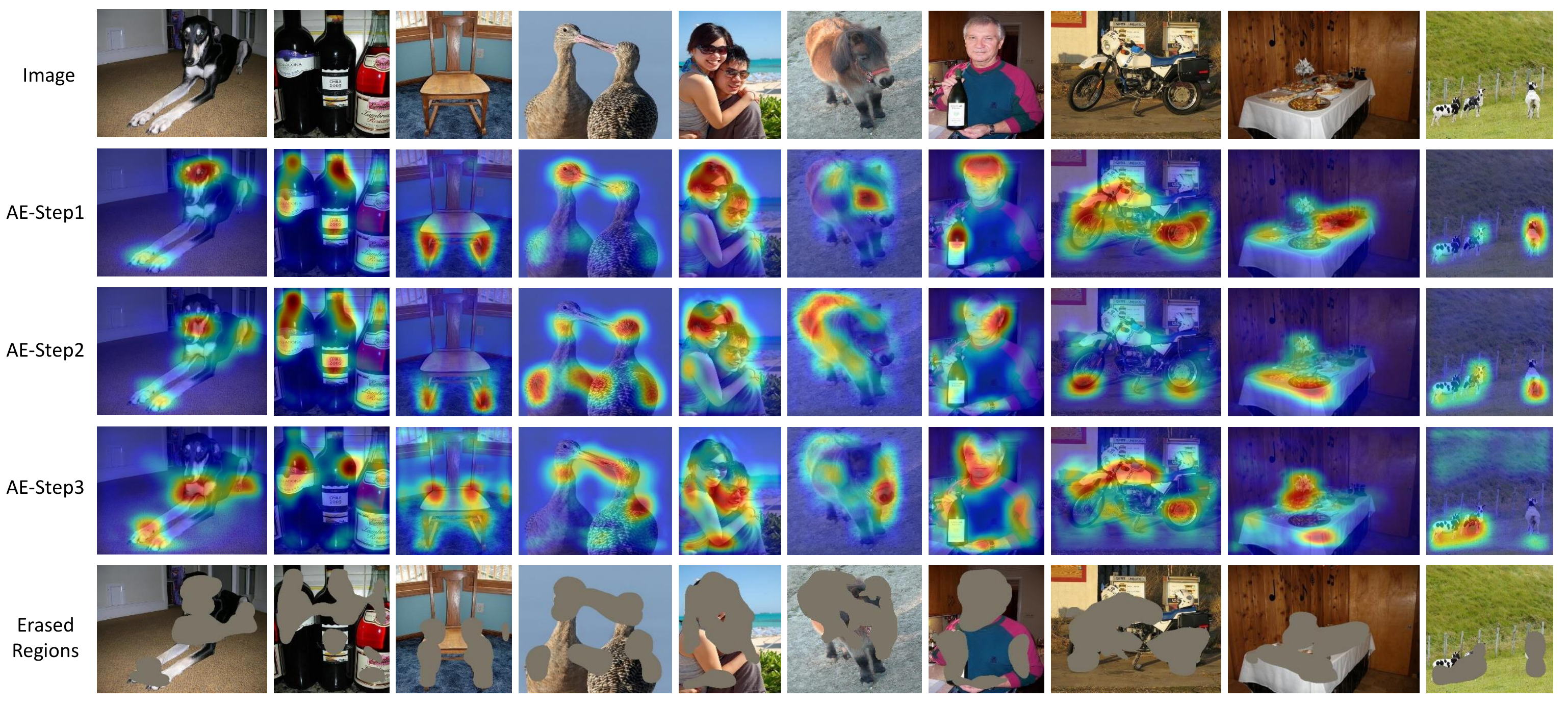}
	\caption{Examples of mined object regions produced by the proposed adversarial erasing approach. The second to fourth rows show the produced heatmaps, where the discriminative regions are highlighted. The images with erased regions are shown in the last row in gray.}
	\label{fig:mining_samples}
	\vspace{-1.5em}
\end{figure*}

For quantitatively understanding the contribution of each AE step, Table~\ref{tab:step} shows the comparison of mIoU scores using foreground regions merged from varying $k\ (k= 1,2,3,4)$ AE steps for training the segmentation network based on DeepLab-CRF-LargeFOV. We can observe that the performance indeed increases as more foreground object regions are added since the segmentation network gets denser supervision. However, after performing four AE steps, the performance drops by 2.1\% due to the \emph{over} \emph{erasing} as explained above. 
Some visualization examples are shown in Figure~\ref{fig:mining_samples}, including training images (top row), heatmaps produced by different AE steps and the finally erased regions (bottom row). We can observe that the AE approach effectively drives the classification network to localize \emph{different} discriminative object regions. For example, regions covering the body of the right-most instance of ``cow" shown in the last column are first localized. By erasing this instance, another two instances on the left side are then discovered. We also conduct experiments on VOC 2012 \emph{test} set using object regions merged from the first three AE steps. The mIoU score is 52.8\%, which outperforms all those methods (as indicated in Table~\ref{tab:test-comp}) only using image-level labels for supervision.

\begin{table*}\setlength{\tabcolsep}{2pt}
	\centering
	\caption{Comparison of segmentation mIoU scores in terms of different training strategies on VOC 2012 \emph{val} set.}
	\label{tab:lgsn}
	\footnotesize
	\begin{tabular}{l|ccccccccccccccccccccc|c}
		\toprule
		Methods  & bkg & plane & bike  & bird  & boat  & bottle & bus   & car   & cat   & chair & cow   & table & dog   & horse & motor & person & plant & sheep & sofa  & train & tv &  mIoU   \\
		\midrule
		w/o PSL & 78.5 & 71.8 & 29.2 & 64.1 & 39.9 & 57.8 & 58.5 & 54.5 & 63.0 & 10.3 & 60.5 & 36.0 & 61.6 & 56.1 & 62.6 & 42.9 & 36.5 & 64.5 & 31.5 & 49.5 & 38.7 & 50.9\\
		w/ PSL & 83.3 & 70.0 & 31.6 & 69.7 & 40.8 & 54.2 & 63.2 & 58.4 & 69.9 & 18.1 & 65.5 & 33.5 & 69.8 & 60.7 & 60.5 & 50.5 & 38.1 & 69.4 & 31.4 & 57.3 & 39.7 & 54.1 \\
		w/ PSL++ & 83.4 & 71.1 & 30.5 & 72.9 & 41.6 & 55.9 & 63.1 & 60.2 & 74.0 & 18.0 & 66.5 & 32.4 & 71.7 & 56.3 & 64.8 & 52.4 & 37.4 & 69.1 & 31.4 & 58.9 & 43.9 & 55.0\\
		\midrule
		w/ PSL+GT、 & 83.6 & 71.0 & 30.6 & 73.0 & 42.7 & 56.1 & 63.6 & 61.7 & 75.2 & 22.2 & 67.6 & 33.4 & 74.6 & 57.8 & 65.6 & 53.6 & 37.7 & 71.6 & 33.2 & 59.0 & 45.1 & 56.1\\
		\bottomrule
	\end{tabular}
	\vspace{-1em}
\end{table*}
\vspace{-1.2em}
\subsubsection{Online PSL for Semantic Segmentation}
\vspace{-0.5em}
We now proceed to evaluate the online PSL and investigate how it benefits the AE approach by discovering auxiliary information. We report the performance of online PSL in Table~\ref{tab:lgsn}, where ``w/o PSL" and ``w/ PSL" denote the result of vanilla DeepLab-CRF-LargeFOV and the proposed PSL method for training, respectively. We can observe that PSL improves the performance by 3.2\% compared with ``w/o PSL", , demonstrating the significant effectiveness of PSL providing additional useful segmentation supervision.

Besides, we perform one more iterative training step on PSL to improve the segmentation results. In particular, we first employ the trained segmentation model from AE and PSL to segment training images. Then, the predicted segmentation masks are used as supervision for training the segmentation network for another round. As shown in Table~\ref{tab:lgsn}, the performance provided by this extra training (denoted as w/ PSL++) is further improved from 54.1\% to 55.0\%. The improvement benefits from the operation of performing CRF on the predicted segmentation masks of training images. After one round training on top of CRF results, the segmentation network has been trained well. We do not observe further performance increase by performing additional training, as no new supervision information is fed in. 

Furthermore, we also examine the effectiveness of our testing strategy where the prohibited threshold is empirically set as 0.1. We utilize ground-truth image-level labels as classification confidences to weight the predicted segmentation score maps (note this is different from the prohibitive information imposed in the training stage). The result is 56.1\% (``w/ PSL + GT"), which is only 1.1\% better than ``w/ PSL ++". Note that ``w/ PSL + GT" actually provides an upper bound on the achievable performance as the score maps are filtered by the ground-truth category annotations and ``w/ PSL ++" performs very closely to this upper bound.

PSL adopts the on-the-fly output of the classification network to re-weight segmentation score maps. Another choice for such classification information is the ground-truth annotation. We also consider the case of using ground-truth image-level labels for prohibiting during the training stage and evaluate the performance. However, using ground-truth information leads to performance drop of 0.6\% compared with our proposed PSL design. This is because PSL effectively exploits the information about object scale that is beneficial for generating more accurate segmentation masks (\ie categories of large objects are preferred with high classification scores compared with those of small objects). Simply using 0-1 ground-truth annotation ignores the scale and performs worse.  We also investigate how PSL performs without using image-level classification confidences and find that the performance drops 1\%. This clearly validates the effectiveness of the proposed online PSL approach using image-level classification information.

\vspace{-0.8em}
\section{Conclusion}
\vspace{-0.5em}
We proposed an adversarial erasing approach to effectively adapt a classification network to progressively discovering and expanding object discriminative regions. The discovered regions are used as pixel-level supervision for training the segmentation network. This approach provides a simple and effective solution to the weakly-supervised segmentation problems. Moreover, we proposed an online prohibitive segmentation learning method, which shows to be effective for mining auxiliary information to AE. Indeed, the PSL method can aid any other weakly-supervised methods. This work paves a new direction of adversarial erasing for achieving weakly-supervised semantic segmentation. In the future, we plan to develop more effective strategies for improving adversarial erasing, such as erasing each training image with adaptive steps or integrating adversarial erasing and PSL into a more unified framework.
\vspace{-0.5em}
\section*{Acknowledgment}
\vspace{-0.5em}
The work is partially supported by the National Key Research and Development of China (No. 2016YFB0800404), National University of Singapore startup grant R-263-000-C08-133, Ministry of Education of Singapore AcRF Tier One grant R-263-000-C21-112 and the National Natural Science Foundation of China (No. 61532005).

{\small
\bibliographystyle{ieee}
\bibliography{egbib}
}

\end{document}